\RequirePackage{amsmath}
\documentclass[runningheads]{llncs}
\usepackage{graphicx}
\usepackage[english]{babel}
\usepackage[utf8x]{inputenc}
\usepackage{amsmath}
\usepackage{subfigure}
\usepackage[colorinlistoftodos]{todonotes}
\usepackage{booktabs}
\usepackage{enumitem}
\usepackage{longtable}
\usepackage{multirow}
\usepackage{adjustbox}
\usepackage{soul,color}
\setlist{nosep, noitemsep}
\title{}

\titlerunning{Chest X-ray Classification Vulnerability Analysis}
\begin{document}
\title{Vulnerability Analysis of Chest X-Ray Image Classification Against Adversarial Attacks}

%\author{Arkadeep Das\inst{1}\and
\author{Saeid Asgari Taghanaki\inst{1}\and
Arkadeep Das\inst{1,2} \and
Ghassan Hamarneh\inst{1}}
\authorrunning{Asgari et al.}
\institute{School of Computing Science, Simon Fraser University, Canada \and{Department of Mathematics, Indian Institute of Technology, Guwahati, India}
\email{\{sasgarit,arkadeep\_das,hamarneh\}@sfu.ca}}
%

%\author{****************************}
%\authorrunning{***}
%\institute{*******************\\
%\email{*****************************}}

\maketitle              % typeset the header of the contribution
\begin{abstract}
Recently, there have been several successful deep learning approaches for automatically classifying chest X-ray images into different disease categories. However, there is not yet a comprehensive vulnerability analysis of these models against the so-called adversarial perturbations/attacks, which makes deep models more trustful in clinical practices. In this paper, we extensively analyzed the performance of two state-of-the-art classification deep networks on chest X-ray images. These two networks were attacked by three different categories (ten methods in total) of adversarial methods (both white- and black-box), namely gradient-based, score-based, and decision-based attacks. Furthermore, we modified the pooling operations in the two classification networks to measure their sensitivities against different attacks, on the specific task of chest X-ray classification.

\keywords{Adversarial perturbation \and chest X-ray classification  \and deep learning}
\end{abstract}
\section{Introduction}
The chest X-ray is among the top most commonly accessible medical imaging examinations used for affordable screening and diagnosis of numerous lung ailments including pneumothorax, mass, cardiomegaly, effusion, and pneumonia.
Owing to huge numbers of patients and increasing burden of lung ailments,  the workload of radiologists has significantly multiplied. Hence, with an intention to accelerate/support the predictions of radiologists, many machine (deep) learning classification frameworks have emerged over the past few years. 

The availability of a new large scale chest X-ray dataset namely "ChestX-ray14"~\cite{2}, which comprises 30,805 patients and 112,120 chest X-ray images, makes it feasible to apply deep learning without a need for data augmentation or synthetic data. Recently, different standard classification deep networks (AlexNet~\cite{24}, VGGNet~\cite{25} and ResNet~\cite{3}) have been applied to this dataset. Wang et al.~\cite{2}  applied pre-trained AlexNet, GoogLeNet~\cite{27}, VGG, and ResNet-50 architectures to classify 8 disease categories. They showed that ResNet-50 achieves superior performance compared to the other applied models. Guendel et al.~\cite{1} proposed  a local aware dense network  for classification of 14 pathology classes in the ChestX-ray14 dataset. Rajpurkar  et al.~\cite{10} proposed CheXNet, a 121-layer convolutional neural network trained on ChestX-ray14 for the pneumonia disease detection task, which exceeds average radiologist performance on the F1 metric. Baltruschat et al.~\cite{11} proposed a fine-tuned ResNet-50 network which achieved high accuracy on 4 out of the 14 disease classes in the chest X-ray dataset. Yao et al.~\cite{12} presented a partial solution to constraints in using LSTMs to leverage inter-dependencies among target labels in predicting 14 pathological classes from chest X-rays. 

The \textit{generalizability} of the deep learning methods, i.e. how they perform on unseen chest X-ray test images, have been explored in the above mentioned works to some extent. However, discovery of "adversarial examples" has exposed serious vulnerabilities in even state-of-the-art deep learning systems~\cite{4}. As of writing there is no comprehensive study on the \textit{vulnerability} analysis of the state-of-the art classification networks against adversarial perturbations for chest X-rays. Samuel G. et al.~\cite{7} considered a single attack, namely projected gradient descent~\cite{8,9} on chest X-ray images.

Adversarial images are crafted by adding perturbations, imperceptible to the naked eye, to the clean images to fool machine learning models. Different categories~\cite{26} of adversarial attacks on images have been recently developed which have been highly successful in fooling deep neural networks. In the medical image analysis domain, attacks may originate during data-transfer through the Internet or local networks~\cite{7}. Even in the case of complete protection from adversarial attacks, training existing deep models with adversarial examples or designing defense mechanisms~\cite{zantedeschi2017efficient} can improve model generalizability and resilience. In this paper, we present a comprehensive analysis of ten different adversarial attacks on classification of chest X-ray images and investigate how two different standard deep neural networks perform against adversarial perturbations. We perform both white (i.e. producing perturbed images using network A and classifying them by the same network) and black-box (i.e. producing perturbed images using network A and classifying them by network B) attacks. 

\section{Methods}

\subsection{Applied deep networks}
We use two state-of-the-art deep models i.e. Inception-ResNet-v2~\cite{5} and Nasnet-Large~\cite{6} to evaluate their performance on classification of both clean and perturbed chest X-ray images. Next, we modify the networks by replacing max-pooling operations with average-pooling to analyze whether the modified networks, especially the ones that are based on single/few pixel perturbation, are less sensitive to attacks. We hypothesize that average-pooling may be more resilient to attacks as it captures more global contextual information from the field of view, instead of selecting a single pixel candidate as max-pooling does.   

\subsection{Applied adversarial attacks}
We applied three different categories of attacks namely gradient-based, score-based, and decision-based:

\begin{itemize}
\item \textbf{Gradient-based} attacks linearize the loss (in our case binary cross-entropy) around an input to which the model predictions for a particular class are most sensitive to. These attacks perturb the image with the gradient of the loss w.r.t. the clean image, gradually and efficiently increasing the magnitude until the model predicts a different label for the perturbed image. 
In our experiments, we have selected five different gradient-based attacks namely,  Fast Gradient Sign Method (G1)~\cite{4}, Projected Gradient Descent (G2)~\cite{8}, DeepFool (G3)~\cite{17}, Linfinity Basic Iterative Method (G4) \cite{4},Limited-memory Broyden–Fletcher–Goldfarb–Shanno Method (L-BFGS) (G5) \cite{19} and we demonstrate how the models trained on clean images perform against the crafted adversarial examples.
\item \textbf{Score-based} attacks rely on confidence scores e.g. softmax class probabilities or logits to numerically estimate the gradient. From this group, we apply Local Search (S1)~\cite{21} (a black box attack based on the greedy local search algorithm to find pixels for which the model is the most sensitive and perturbing them to misclassify the input) and the Single Pixel (S2)~\cite{20} attacks. 
\item \textbf{Decision-based} attacks~\cite{22} solely rely on the predicted class or label of the model without requiring gradients or logits. From this group, we applied Gaussian Blur (D1), Contrast Reduction (D2) and Additive Gaussian Noise (D3) in our experiments. In all of the aforementioned attacks, a line-search is performed internally to find minimal perturbations required by the image to turn it into an adversarial example.
\end{itemize}

We trained both the networks from scratch with a batch size of 32 and 8 for training the Inception-ResNet-v2 and Nasnet-Large, respectively. RMSProp optimizer \cite{16} with a decay of 0.9 and $\epsilon = 1$ and an initial learning rate of 0.045, decayed every 4 epochs using an exponential rate of 0.94 were used for all of our experiments as described in~\cite{5,6}. We set all attack parameters as proposed by their authors and utilized Foolbox~\cite{23}, to craft adversarial examples.

\section{Dataset}
We use ChestX-ray14 dataset~\cite{2} which comprises 112,120 gray-scale images with 14 disease labels and 1 no-finding label. We treat all the disease classes as positive and formulate a binary classification task of "disease" vs. "non-disease". We randomly selected 95,128 images for training and 16,792 for validation. We randomly picked 200 unseen images as the test set, with 93 images with chest disease labels and 107 having "no finding" labels. These clean images are used for carrying out different adversarial attacks and the models trained on clean images are evaluated against them.  

\section{Results and discussion}
\begin{figure}
  \centering
    \includegraphics[width=1\textwidth]{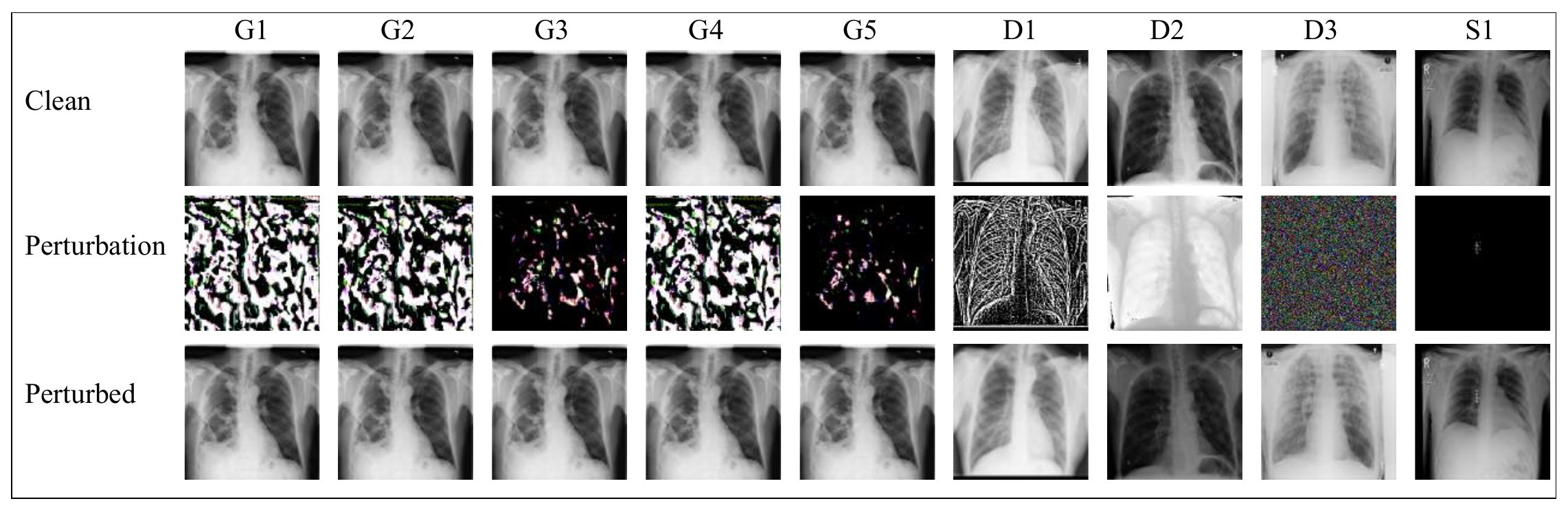}
    \caption{Perturbed images produced by 10 (3 categories) different attacks.}
    \label{fig1}
\end{figure}

\begin{figure}
  \centering
    \includegraphics[width=.8\textwidth]{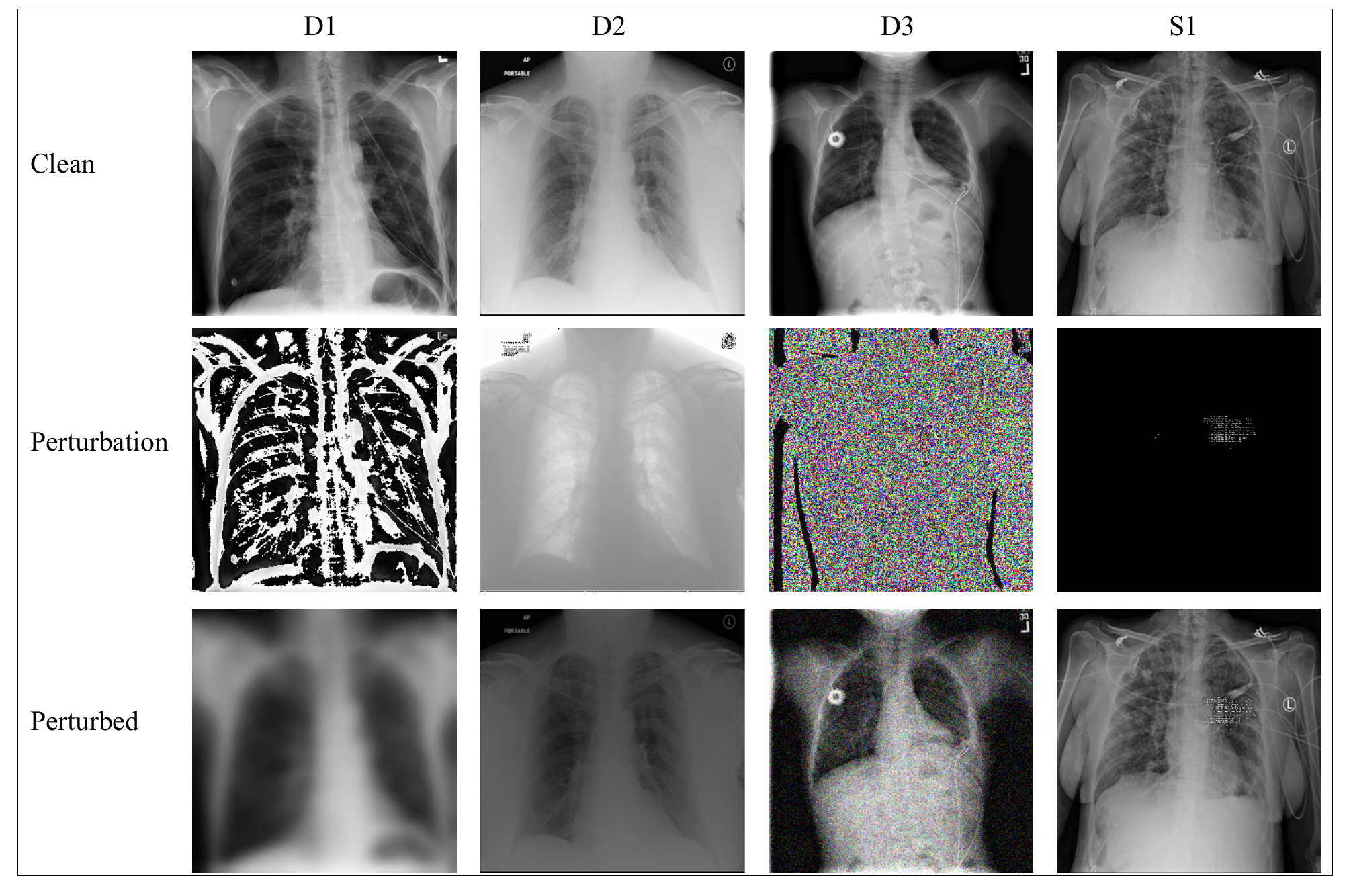}
    \caption{Human perceptible adversarial perturbations}
    \label{fig2_2}
\end{figure}

Figure~\ref{fig1} shows the perturbed images produced by the ten different applied attacks. In Figure~\ref{fig2_2}, we visualize a few samples where the perturbations are perceptible by human. We observed that most of the produced images by D1 (i.e Gaussian blur), D2 (i.e. contrast reduction), D3 (i.e  additive Gaussian noise), S1 (i.e. local search) attack can be easily detected by the naked eye. We also found that S1 requires relatively more time compared to other methods to find an adversarial image.

In Tables~\ref{table1} and~\ref{table2}, we report accuracy and area under ROC for two networks with/without modification for clean and ten different adversarial attacks (white- and black-box). Note that the single pixel attack~\cite{20} i.e. S2 (from the score based attacks category) failed to fool the networks for the entire test set which shows the single pixel attack works well on RGB (colored images) but not on gray-scale X-ray images as it is not simple to fool a deep model by changing only a single "gray-scale" pixel. 

\begin{table}
\small
\caption{Performance of original/modified Inception-ResNet-v2 (IR2) and Nasnet-Large (NL) against ten different \textit{white-box} attacks. In the Table, MP, AP, Acc., AU refer to max-pooling, average-pooling, accuracy, and area under ROC, respectively.}
\label{table1}
\centering
\setlength{\tabcolsep}{3pt}
\begin{tabular}{lllccccccccccc}
\hline
 & \multirow{2}{*}{Model} & \multicolumn{1}{c}{\multirow{2}{*}{Metrics}} & \multirow{2}{*}{Clean} & \multicolumn{5}{c}{Gradient} & \multicolumn{3}{c}{Decision} & \multicolumn{2}{c}{Score} \\ \cline{5-14} 
 &  & \multicolumn{1}{c}{} &  & G1 & G2 & G3 & G4 & G5 & D1 & D2 & D3 & S1 & S2\\ \hline
\multirow{4}{*}{MP} & \multirow{2}{*}{IR2} & Acc. & 0.70 & 0.00 & 0.00 & 0.00 & 0.00 & 0.00 & 0.04 & 0.10 & 0.32 & 0.65 & 0.70 \\
 &  & AU & 0.75 & 0.00 & 0.00 & 0.00 & 0.00 & 0.00 & 0.06 & 0.19 & 0.52 & 0.74 & 0.75 \\
 & \multirow{2}{*}{NL} & Acc. & 0.73 & 0.00 & 0.00 & 0.01 & 0.00 & 0.00 & 0.06 & 0.41 & 0.30 & 0.32 & 0.73 \\
 &  & AU & 0.77 & 0.00 & 0.00 & 0.10 & 0.00 & 0.00 & 0.10 & 0.66 & 0.58 & 0.55 & 0.77\\ \hline
\multirow{4}{*}{AP} & \multirow{2}{*}{IR2} & Acc. & 0.71 & 0.00 & 0.00 & 0.00 & 0.00 & 0.00 & 0.04 & 0.24 & 0.14 & 0.62 & 0.71 \\
 &  & AU & 0.74 & 0.00 & 0.00 & 0.00 & 0.00 & 0.00 & 0.06 & 0.39 & 0.26 & 0.72 & 0.74 \\
 & \multirow{2}{*}{NL} & Acc. & 0.72 & 0.00 & 0.00 & 0.00 & 0.00 & 0.00 & 0.03 & 0.41 & 0.48 & 0.72 & 0.72\\
 &  & AU & 0.74 & 0.00 & 0.00 & 0.00 & 0.00 & 0.00 & 0.04 & 0.64 & 0.64 & 0.74 & 0.74\\ \hline
\end{tabular}
%\end{adjustbox}
\end{table}

\begin{table}
\small
\caption{Performance of original/modified Inception-ResNet-v2 (IR2) and Nasnet-Large (NL) against ten different \textit{black-box} attacks. In the Table, MP, AP, Acc., AU refer to max-pooling, average-pooling, accuracy, and area under ROC, respectively.}
\label{table2}
\centering
\setlength{\tabcolsep}{3pt}
\begin{tabular}{lllccccccccccc}
\hline
 & \multirow{2}{*}{Model} & \multicolumn{1}{c}{\multirow{2}{*}{Metrics}} & \multirow{2}{*}{Clean} & \multicolumn{5}{c}{Gradient} & \multicolumn{3}{c}{Decision} &\multicolumn{2}{c}{Score} \\ \cline{5-14} 
 &  & \multicolumn{1}{c}{} &  & G1 & G2 & G3 & G4 & G5 & D1 & D2 & D3 & S1 & S2 \\ \hline
\multirow{4}{*}{MP} & \multirow{2}{*}{IR2} & Acc. & 0.70 & 0.46 & 0.43 & 0.43 & 0.43 & 0.43 & 0.53 & 0.81 & 0.36 & 0.45 & 0.70\\
 &  & AU & 0.75 & 0.44 & 0.41 & 0.40 & 0.41 & 0.41 & 0.43 & 0.84 & 0.24 & 0.40 & 0.75 \\
 & \multirow{2}{*}{NL} & Acc. & 0.73 & 0.53 & 0.51 & 0.51 & 0.51 & 0.51 & 0.57 & 0.58 & 0.74 & 0.56 & 0.73\\
 &  & AU & 0.77 & 0.52 & 0.49 & 0.49 & 0.49 & 0.49 & 0.51 & 0.55 & 0.82 & 0.55 & 0.77 \\ \hline
\multirow{4}{*}{AP} & \multirow{2}{*}{IR2} & Acc. & 0.71 & 0.51 & 0.52 & 0.52 & 0.52 & 0.51 & 0.53 & 0.29 & 0.40 & 0.53  & 0.71\\
 &  & AU & 0.74 & 0.49 & 0.49 & 0.49 & 0.47 & 0.50 & 0.47 & 0.24 & 0.40 & 0.52 & 0.74 \\
 & \multirow{2}{*}{NL} & Acc. & 0.72 & 0.59 & 0.58 & 0.58 & 0.58 & 0.58 & 0.49 & 0.53 & 0.51 & 0.38  & 0.72\\
 &  & AU & 0.74 & 0.59 & 0.58 & 0.58 & 0.58 & 0.58 & 0.46 & 0.52 & 0.46 & 0.39 & 0.74\\ \hline
\end{tabular}
%\end{adjustbox}
\end{table}

As reported in Table~\ref{table1}, the gradient based attacks were almost completely successful in fooling both networks (with/without modification) when the victim model for attack was the same reference model, i.e. in a white-box attack scenario. The decision and score based attacks were almost unsuccessful in fooling the models. We observed that Nasnet-Large with average pooling was 18\% stronger in comparison to Nasnet-Large with max pooling. Note that the local search attack (S1) completely failed against Nasnet-Large with average-pooling.

\begin{figure}[h!]
\centering     %%% not \center
\subfigure[Nasnet-Large]{\label{fig:a}\includegraphics[width=60mm]{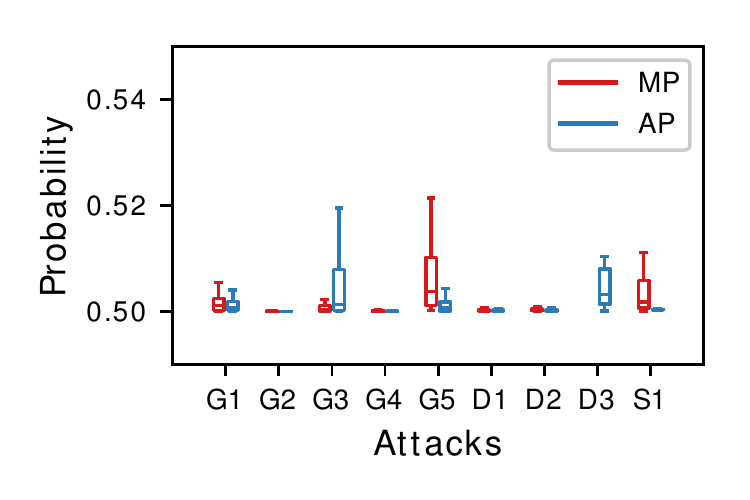}}
\subfigure[Inception-ResNet-v2]{\label{fig:b}\includegraphics[width=60mm]{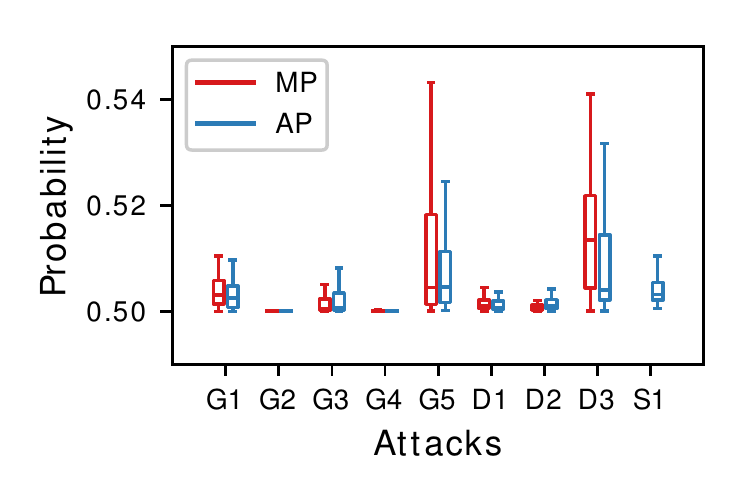}}
\caption{Probability values of the original and modified networks after attack}
\label{fig2}
\end{figure}

In Table~\ref{table2}, we show the performance of both the networks against the black-box attacks i.e. we craft adversarial images with Inception-ResNet-v2, but test them with Nasnet-Large and vice versa. As reported in the table, almost all the methods were partially successful but not as high as white-box attacks. For gradient based black-box attacks, average pooling shows more resiliency against the attacks. We observed that for $23\%\pm9\%$ and $27\%\pm8\%$ of the test samples both the networks failed on the same cases for average and max pooling, respectively.

In Figure~\ref{fig2}, we show the probability values of the two networks (with/without modification) only after successful attacks on the disease class. Higher ranges/values indicate a stronger attack or a more vulnerable network. As shown in the figure, D3 (i.e. Additive Gaussian Noise), S1 (i.e. Local search) and G5 (i.e. L-BFGS) attacks are highly sensitive to the choice of pooling (max/average) operation. The range of the attack's confidence varies from $\sim0.50$ to $\sim0.55$. Note that absence of a box in the figure means there was no successful attack for a disease class in that experiment. 
\begin{figure}
  \label{fig4}
  \centering
    \includegraphics[width=.6\textwidth]{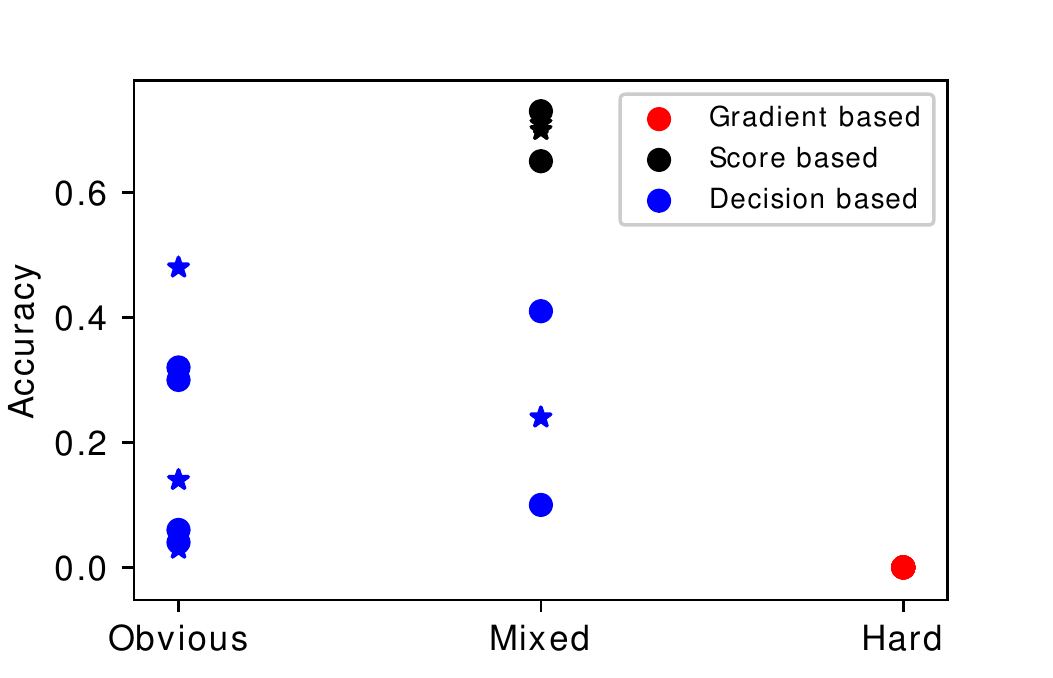}
    \caption{Distribution of adversarial images crafted based on Inception-ResNet-v2 network w.r.t. human perception. The dots and stars in the figure refer to max and average pooling, respectively. The words Obvious, Mixed, and Hard refer to the level of difficulty for a human to perceive the attacks.}
\end{figure}
In Figure~\ref{fig4}, we visualize the accuracy of Inception-ResNet-v2 for different groups of attacks and, in the same plot, we show the perceptibility of each group of the attacks (i.e. the difficulty level for a human to detect a perturbed image). Note that lower accuracy and harder detection (lower right of the plot) implies a more successful attack. As shown in the figure, gradient based attacks are the most successful ones in terms of fooling both human (i.e. perception) and machine (i.e. accuracy).

\section{Conclusion}
In this paper, we extensively tested the vulnerability of the two state-of-the-art deep classification networks against ten different adversarial attacks on chest X-ray images. We found that the single pixel attack completely failed for gray-level X-ray chest images. We also showed that the pooling operation can make a considerable difference for some attacks, even leading to a complete failure of the attack for a particular class. We also demonstrated that the crafted adversarial images with some of the attacks, e.g. Gaussian blur and contrast reduction methods, can be simply detected with the naked eye. Finally, we showed that the gradient based attacks applied to the chest X-ray images are the most successful in terms of fulling both machine and human. Although both networks, Inception-ResNet-v2 and Nasnet-Large, failed against gradient-based attacks, in general, the latter (with average pooling) was more resilient to decision and score based attacks.

\section*{Acknowledgments}
We thank NVIDIA Corporation for GPU donation and MITACS Globalink for funding.
\bibliographystyle{splncs04}
\bibliography{sample.bib}

\end{document}